\documentclass[runningheads]{llncs}
\usepackage[T1]{fontenc}
\usepackage{graphicx}
\usepackage{amsmath,amssymb}
\usepackage{booktabs}
\usepackage{multirow}
\usepackage[table]{xcolor}
\usepackage{microtype}
\usepackage{orcidlink}
\newif\ifanonymous
\anonymousfalse
\begin{document}
\title{DSSR-3D: Decoupled Reasoning for View-Dependent Referring in 3D Gaussians}
\titlerunning{DSSR-3D}
%
\ifanonymous
  \author{Anonymous Author(s)}
  \authorrunning{Anonymous Author(s)}
  \institute{Anonymous Institution}
\else
  \author{Thanh-Khoi Nguyen\inst{1,2}\orcidlink{0009-0003-5879-451X} \and
  Thien-Phuc Tran\inst{1,2}\orcidlink{0009-0008-0800-6884} \and
  Minh-Triet Tran\inst{1,2}\orcidlink{0000-0003-3046-3041}\thanks{Corresponding author.}}
  \authorrunning{T.-K. Nguyen et al.}
  \institute{University of Science, Ho Chi Minh City, Vietnam \and
  Viet Nam National University, Ho Chi Minh City, Vietnam\\
  \email{23120009@student.hcmus.edu.vn, ttphuc@selab.hcmus.edu.vn, tmtriet@fit.hcmus.edu.vn}}
\fi
\maketitle              
\begin{abstract}
Recent advances in 3D Gaussian Splatting have enabled open-\allowbreak{}vocabulary and referring segmentation by distilling semantic knowledge from 2D foundation models into 3D representations. However, existing referring fields embed language features in a globally view-invariant space, making them fundamentally unable to resolve observer-centric spatial relations (e.g., "to the left of") that depend on camera pose. We propose DSSR-3D, an inference-time framework for view-dependent referring segmentation on continuous 3D Gaussian fields, formalized as two interfaces - pose-invariant semantic localization and pose-conditioned spatial reasoning - such that any pair of functions satisfying these constraints yields a valid instantiation, requiring no retraining of the underlying semantic field and no reliance on discrete geometric proxies such as bounding boxes. We instantiate the two interfaces with a temperature-sharpened softmax localization mechanism and a projection-based directional scoring function, fused via a lightweight, training-free step, and show they transfer zero-shot to structurally distinct semantic fields without adaptation. We further propose ViewRef-GS, a benchmark isolating view-dependent segmentation on 3D Gaussian fields, evaluated jointly with an augmented Ref-LERF to provide a comprehensive testbed for viewpoint-dependent spatial grounding. Experiments show consistent gains over existing 3DGS-based referring methods, with no additional training beyond the base semantic field.

\end{abstract}

\section{Introduction}
\label{sec:intro}

\begin{figure}[!t]
    \centering
    \includegraphics[width=0.5\linewidth]{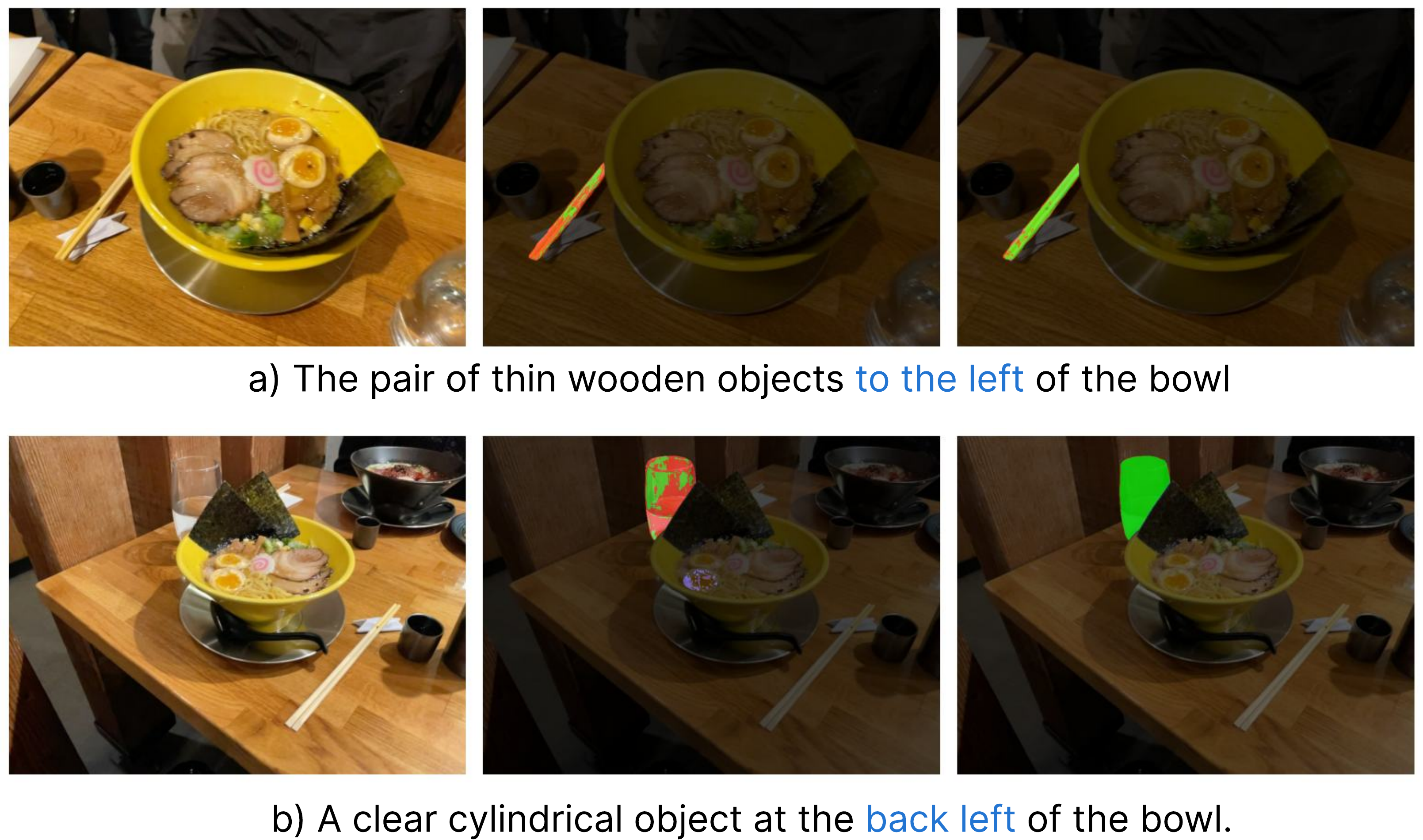}
    \caption{Failure cases under viewpoint ambiguity and occlusion. (a) View-invariant fields fail on observer-centric queries. (b) 2D foundation models localize visible regions instead of occluded targets. (Green: True positive, Red: False negative)}
    \label{fig:motivation}
\end{figure}

Beyond novel view synthesis~\cite{kerbl20233dgs,mildenhall2020nerf}, 3D Gaussian Splatting (3DGS) has enabled 3D segmentation by distilling 2D foundation models (CLIP~\cite{radford2021clip}, SAM~\cite{kirillov2023sam}) into queryable semantic fields~\cite{qin2024langsplat,zhou2024feature3dgs,ye2024gaussiangrouping}, recently extended to natural-language referring segmentation by ReferSplat~\cite{he2025refersplat}. Yet these methods~\cite{qin2024langsplat,he2025refersplat,zhou2024feature3dgs,zhao2024fmgs,shi2024language,ji2025fastlgs} encode semantics in globally view-invariant spaces without camera-pose conditioning, making observer-centric spatial relations difficult to resolve, leading to severe spatial hallucinations and physical inconsistencies in dynamic interactive scenarios. The problem is compounded on the 2D side: foundation models such as GroundingSAM~\cite{ren2024grounded} operate directly on isolated image frames without any notion of depth or 3D structure, so their spatial reasoning breaks down under occlusion and view inconsistency (Fig.~\ref{fig:motivation}). Recent 3D Scene Graph methods~\cite{tao2026geocga,xie2026relags} model spatial relations (e.g., ``above'') explicitly, but as static, view-invariant edges, remaining ill-suited to relations whose meaning shifts with viewpoint.

To address perspective-dependent queries, pioneering works explicitly incorporate observer viewpoints, either by conditioning 3D multimodal models on camera poses~\cite{nanri2026viewpoint3dseg} or training dedicated viewpoint prediction networks for scene alignment~\cite{shi2024viewpointaware}. However, these approaches face three critical bottlenecks. First, they operate on traditional point clouds, omitting the real-time rendering advantages of 3DGS. Second, they incur substantial training overhead, as camera dependence is implicitly learned through additional viewpoint supervision and pretraining, rather than being directly available for inference-time evaluation. Finally, they rely heavily on discrete 3D bounding boxes. Gathering dense box annotations is prohibitively expensive, and forcing spatial reasoning through such rigid proxies introduces quantization errors that affect fine-grained geometric fidelity. This leaves a critical gap: how can we achieve true viewpoint-aware spatial reasoning natively on 3D Gaussian fields without relying on point clouds, discrete bounding boxes, or heavy training overhead?

To address these limitations, we propose a verifiable, training-free reasoning layer that decouples spatial reasoning to achieve interpretable, evidence-grounded spatial grounding natively on continuous 3DGS. Rather than costly bounding-box annotations or scale-sensitive, noise-prone fixed top-$K$ selection, we propose a Temperature-Sharpened Softmax mechanism for scale-adaptive, bounding-box-free localization, paired with a Direction-Aware Scoring mechanism that projects Gaussians onto the observer's local horizontal $XZ$-plane with a soft spatial decay to inject directional bias, without task-specific retraining. Both instantiate a single principle: we formalize view-dependent spatial grounding as two independent, interface-constrained sub-problems (Sec.~\ref{sec:problem-formulation}), decoupling spatial reasoning from the underlying semantic field and enabling zero-shot composition with structurally distinct frozen fields (Table~\ref{tab:cross-backbone}). Finally, we propose \textbf{ViewRef-GS}, to our knowledge the first benchmark isolating view-dependent, observer-centric spatial relations for 3DGS referring segmentation, combined with a directional-query augmentation of Ref-LERF into a six-scene evaluation suite. Our main contributions are as follows:
\begin{itemize}
    \item We formalize view-dependent 3D spatial grounding as two independent sub-problems: pose-invariant localization and pose-conditioned relational scoring. This training-free decomposition acts as a plug-and-play reasoning layer, achieving consistent gains across structurally distinct, frozen 3DGS backbones without per-backbone tuning.
    \item We instantiate this via two inference-time mechanisms: a Temperature-Sharpened Softmax for scale-adaptive, bounding-box-free reference localization, and a Direction-Aware Scoring function that resolves pose-conditioned relations through projection and spatial decay.
    \item We propose ViewRef-GS, a synthetic benchmark that, together with an augmented Ref-LERF, provides a comprehensive testbed for evaluating viewpoint-dependent spatial grounding under diverse observer-centric configurations.
\end{itemize}

\section{Related Work}
\label{sec:related-work}
\subsection{3D Semantic and Referring Segmentation}
\label{sec:related-semantic-segmentation}
3DGS~\cite{kerbl20233dgs} has become a dominant explicit representation for real-time novel-view synthesis, extending earlier neural radiance fields (NeRF)~\cite{mildenhall2020nerf}. A large body of work embeds semantic or language features into Gaussians: LERF~\cite{kerr2023lerf} first volume-rendered CLIP embeddings inside NeRF, and subsequent 3DGS methods-Feature3DGS~\cite{zhou2024feature3dgs}, LangSplat~\cite{qin2024langsplat}, LEGaussian~\cite{shi2024language}, FMGS~\cite{zhao2024fmgs}, FastLGS~\cite{ji2025fastlgs}-distill CLIP/SAM features per-Gaussian for open-vocabulary querying, while GaussianGrouping~\cite{ye2024gaussiangrouping} combines SAM masks with contrastive grouping for instance-level segments.
ReferSplat~\cite{he2025refersplat} formalized referring segmentation on 3DGS via spatially-aware referring fields and the Ref-LERF benchmark. Successors enrich relational modeling through cross-modal graph alignment or explicit relational embeddings (GeoCGA~\cite{tao2026geocga}, ReLaGS~\cite{xie2026relags}), or inject camera geometry with cross-view logit alignment (CaRF~\cite{tao2025carf}). Yet all encode semantics in a globally view-invariant space, limiting their ability to resolve observer-centric spatial relations whose meaning depends on camera pose.

\subsection{Viewpoint-Aware Spatial Reasoning}
\label{sec:related-viewpoint-reasoning}
Spatial language in 3D is inherently viewpoint-dependent, with relations such as "left/right" or "front/behind" shifting with the observer's pose. Early work conditions grounding on the predicted speaker viewpoint~\cite{shi2024viewpointaware}, and Nanri et al.~\cite{nanri2026viewpoint3dseg} introduce a large-scale benchmark that isolates observer-centric relations under camera-pose conditioning. Subsequent methods target multi-view consistency and relational reasoning-ViewSRD~\cite{huang2025viewsrd} decomposes multi-anchor queries across viewpoints, Position3D~\cite{wang2026position3d} injects spatial relations into point-cloud queries, and others explore relationship graphs, zero-shot view retrieval, or VLM-guided novel-view selection on 3DGS~\cite{shi2025semanticsprogramming,liao2025gvr,mvggt2026,kim2026gaussexplorer}-but all either operate on point clouds/\allowbreak{}meshes, rely on discrete bounding boxes, require viewpoint-specific training, or treat relations as view-invariant. DSSR-3D instead natively targets continuous 3D Gaussian fields with two training-free interfaces-pose-invariant localization and pose-conditioned directional scoring-requiring neither boxes nor per-backbone retraining, accompanied by ViewRef-GS, the first benchmark isolating view-dependent referring segmentation on 3DGS.

\section{Method}
\label{sec:method}
\subsection{Preliminary: 3D Gaussian Splatting}
\label{preliminary}
A scene is represented as Gaussians $\mathcal{G} = \{g_i\}_{i=1}^{\mathcal{N}}$, each parameterized by mean $\mu_i \in \mathbb{R}^3$, covariance $\Sigma_i \in \mathbb{R}^{3\times3}$, and opacity $\alpha_i \in [0,1]$; for novel view synthesis, each additionally encodes view-dependent color $c_i$ via spherical harmonics. Rendering projects Gaussians onto the image plane via depth-sorted alpha compositing (Eq.~\ref{eq:color-rendering}). For referring segmentation, we associate each Gaussian with a semantic feature $S_i$ and replace $c_i$ with a scalar alignment score $m_i$, yielding the query-conditioned response map $M(V)$ (Eq.~\ref{eq:score-rendering}), where $m_i$ measures $g_i$'s relevance to the text query.

\vspace{-0.30em}
\noindent
\begin{minipage}{0.49\linewidth}
\begin{equation}
\resizebox{0.82\linewidth}{!}{$\displaystyle
C(V) = \sum_{i=1}^{\mathcal{N}} c_i \alpha_i \prod_{j=1}^{i-1} (1 - \alpha_j)
$}
\label{eq:color-rendering}
\end{equation}
\end{minipage}%
\hfill
\begin{minipage}{0.49\linewidth}
\begin{equation}
\resizebox{0.82\linewidth}{!}{$\displaystyle
M(V) = \sum_{i=1}^{\mathcal{N}} m_i \alpha_i \prod_{j=1}^{i-1} (1 - \alpha_j)
$}
\label{eq:score-rendering}
\end{equation}
\end{minipage}
\vspace{-0.30em}

\subsection{Problem Formulation}
\label{sec:problem-formulation}
Let $G = \{g_i\}_{i=1}^{N}$ be a 3D Gaussian scene. Given a camera viewpoint $V$ and a query $T$, we aim to localize the target object. We focus on observer-centric, view-dependent relations (e.g., "left", "front") and isolate viewpoint ambiguity by projecting geometry onto the camera's local horizontal $XZ$ plane. Motivated by ground-based robotics, we constrain $V$ to near-planar cameras (excluding aerial views where lateral constraints become ill-defined) and focus on single-anchor queries. Prior work conflates semantic identification and spatial reasoning into a single scoring function, forcing retraining per semantic field. We decompose the problem into two independent sub-problems, applied atop the frozen field's native scoring $\{S_i\}$ of the anchor or target span:
\begin{itemize}
    \item \emph{Problem I (View-Invariant Localization):} Given $\{S_i^{\mathrm{anchor}}\}$, estimate a pose-invariant reference $r \in \mathbb{R}^3$ that is \emph{scale-adaptive} and \emph{bounded} within the anchor's semantic support.
    \item \emph{Problem II (View-Conditioned Relational Scoring):} Given $r$, pose $V$, and directional constraint $d$, estimate a spatial score $s_i^{\mathrm{dir}} \in [-1,1]$ that is \emph{pose-conditioned}, \emph{sign-consistent}, and \emph{locality-aware} (decaying with distance from $r$).
\end{itemize}
The target's score $s_i^{\mathrm{sem}}$ uses the same scoring on $T_{\mathrm{target}}$ and is fused with $s_i^{\mathrm{dir}}$ in Sec.~\ref{sec:semantic-spatial-fusion}, not a third interface. Any functions $L: \{S_i^{\mathrm{anchor}}\} \to r$ and $R: (\mu_i, r, V, d) \to s_i^{\mathrm{dir}}$ satisfying these constraints are valid instantiations. Because $r$ is invariant to $V$, all camera dependence is isolated in Problem II; and since neither $L$ nor $R$ depends on how $\{S_i\}$ was produced, both transfer across different semantic fields without retraining (Table~\ref{tab:cross-backbone}).

\subsection{Pipeline Overview}
\label{sec:pipeline-overview}

\begin{figure}[!t]
    \centering
    \includegraphics[width=1.0\linewidth]{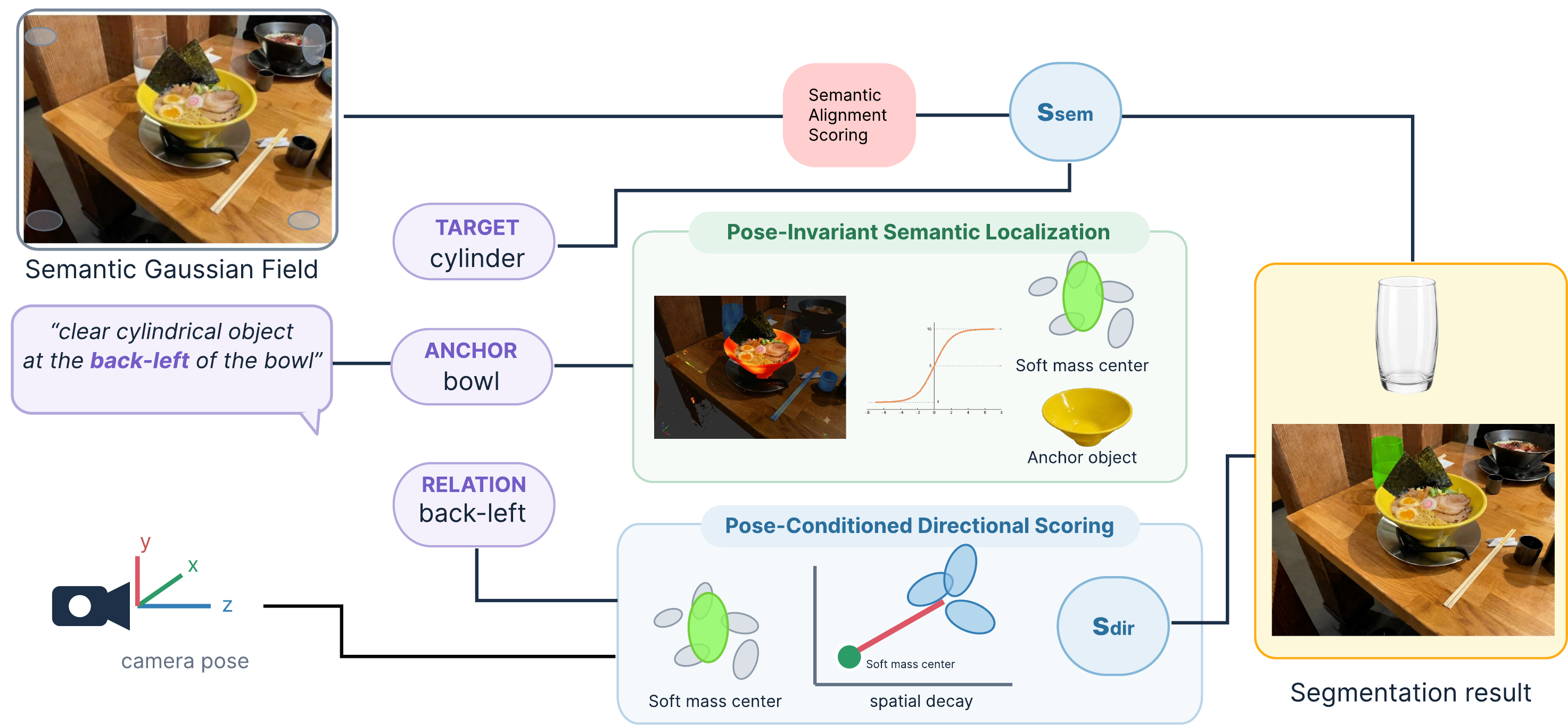}
    \caption{Overview of the DSSR-3D pipeline.}
    \label{fig:pipeline}
\end{figure}

Our training-free pipeline maps a query $T$ to a 2D mask over a frozen semantic field (Figure~\ref{fig:pipeline}). Query Rewriting extracts $(T_{\mathrm{target}}, T_{\mathrm{dir}}, T_{\mathrm{anchor}})$; \textbf{Pose-Invariant Semantic Localization} (Sec.~\ref{sec:anchor-localization}) estimates the reference $\mathbf{r}$; \textbf{Pose-Conditioned Directional Scoring} (Sec.~\ref{sec:direction-scoring}) evaluates the relation under camera pose $V$; and \textbf{Semantic-Spatial Fusion} (Sec.~\ref{sec:semantic-spatial-fusion}) renders the final mask. Isolating camera dependence in directional scoring keeps the pipeline plug-and-play across frozen semantic fields.

\subsection{Pose-Invariant Semantic Localization}
\label{sec:anchor-localization}

To instantiate Problem I, localization interface $L$ must distill a continuous, noisy semantic field into a single stable coordinate $\mathbf{r}$ that faithfully reflects an object's true spatial extent. We parse queries into $(T_{\mathrm{target}}, T_{\mathrm{dir}}, T_{\mathrm{anchor}})$ via FLAN-T5-Small~\cite{chung2024scaling}, and evaluate $T_{\mathrm{anchor}}$ against the frozen, externally supplied Gaussian field to yield a semantic confidence score $S_i^{\mathrm{anchor}}$ for each Gaussian. 

Treating localization as a hard selection problem (e.g., averaging the top-$K$ highest-scoring Gaussians) collapses a continuous signal into a brittle binary decision. Because object scale varies by orders of magnitude, a fixed cardinality $K$ inevitably discards genuine semantic support for large objects or absorbs background noise for compact ones, failing to resolve both extremes simultaneously (Table~\ref{tab:ablation_center_of_mass}). 

Rather than relying on fixed cardinality, we formulate localization as a soft aggregation. We define a dynamic candidate set $\mathcal{K} = \{i \mid S_i^{\mathrm{anchor}} > S_{th}\}$ using a base confidence floor $S_{th}$ to exclude non-diagnostic noise. To allow each Gaussian's geometric contribution to scale continuously with its confidence, we apply a temperature-scaled softmax weighting:
\par\noindent
\vspace{0.35em}
\makebox[\linewidth][c]{%
\begin{minipage}{0.49\linewidth}
\begin{equation}
\scalebox{0.82}{$\displaystyle
w_i = \frac{\exp(S_i^{\mathrm{anchor}} / \tau)}{\sum_{j \in \mathcal{K}} \exp(S_j^{\mathrm{anchor}} / \tau)},
$}
\label{eq:anchor-softmax-weight}
\end{equation}
\end{minipage}%
\hfill
\begin{minipage}{0.49\linewidth}
\begin{equation}
\scalebox{0.82}{$\displaystyle
    \mathbf{r} = \sum_{i \in \mathcal{K}} w_i \mu_i.
    \vphantom{\frac{\exp(S_i^{\mathrm{anchor}} / \tau)}{\sum_{j \in \mathcal{K}} \exp(S_j^{\mathrm{anchor}} / \tau)}}
$}
\label{eq:anchor-centroid}
\end{equation}
\end{minipage}
}
where $\tau$ calibrates distributional sharpness independently of object geometry. The reference coordinate is then analytically derived as the weighted centroid:


Crucially, because $\mathbf{r}$ is a convex combination of Gaussian means, it is mathematically guaranteed to remain within the semantic support's convex hull. By structurally accommodating scale variance and explicitly bounding the output to this hull, this analytical formulation satisfies the \emph{scale-adaptive} and \emph{bounded} constraints of Problem I, independent of the underlying semantic field (Table~\ref{tab:cross-backbone}).

\vspace{-2mm}
\subsection{Pose-Conditioned Directional Scoring}
\label{sec:direction-scoring}
\vspace{-2mm}

To instantiate Problem II, the relational interface $R$ maps the pose-invariant reference $\mathbf{r}$ and camera pose $V = (K_{\mathrm{int}}, E)$ into a view-dependent spatial score $s_i^{\mathrm{dir}}$. To function entirely at inference time, this mapping is designed to satisfy three constraints: it must be \emph{pose-conditioned}, \emph{locality-aware}, and \emph{bounded} with sign-consistency. 

We address the \emph{pose-conditioned} requirement by transforming both the Gaussian means $\mu_i$ and the reference $\mathbf{r}$ into the camera coordinate frame via the extrinsic matrix $E$. To isolate the viewpoint, we compute the relative offset in the camera's local horizontal $XZ$-plane:
\par\noindent
\vspace{0.35em}
\makebox[\linewidth][c]{%
\begin{minipage}{0.49\linewidth}
\begin{equation}
\scalebox{0.88}{$\displaystyle
    \Delta x_i = (E\mu_i)_{xz} - (E\mathbf{r})_{xz}.
    \vphantom{\frac{p^{\mathrm{raw}}_i}{1 + \|\Delta x_i\| / \sigma_d}}
$}
\label{eq:camera-plane-offset}
\end{equation}
\end{minipage}%
\hfill
\begin{minipage}{0.49\linewidth}
\begin{equation}
\scalebox{0.88}{$\displaystyle
p^{\mathrm{pen}}_i = \frac{p^{\mathrm{raw}}_i}{1 + \|\Delta x_i\| / \sigma_d},
$}
\label{eq:spatial-decay}
\end{equation}
\end{minipage}
}

Given a normalized directional vector $d \in \mathbb{R}^2$ parsed from the query, the raw directional alignment is computed via algebraic projection: $p^{\mathrm{raw}}_i = \Delta x_i \cdot d$.

While this projection establishes correct geometric orientation, relying on it alone risks favoring distant objects that are merely geometrically aligned. To implement the \emph{locality-aware} constraint, we apply a soft spatial decay factor that prioritizes the local neighborhood of the anchor:
where $\sigma_d$ is an adaptive spatial scale determined by the median distance of top semantic candidates from the anchor. This step suppresses artifacts that are geometrically aligned but contextually too distant.

Finally, because $p^{\mathrm{pen}}_i$ remains unbounded, it cannot be directly fused with semantic probabilities. We satisfy the \emph{bounded} and \emph{sign-consistent} constraints by passing the penalized score through a hyperbolic tangent transformation:
\begin{equation}
    s^{\mathrm{dir}}_i = \tanh\left( \frac{p^{\mathrm{pen}}_i}{\Sigma_{\mathrm{proj}} + \epsilon} \right),
\label{eq:directional-score}
\end{equation}
where $\Sigma_{\mathrm{proj}}$ acts as a scene-wide normalization factor and $\epsilon$ ensures numerical stability. This mapping bounds the output to the $[-1, 1]$ range while preserving the geometric sign (e.g., positive for the queried direction, negative for the opposite).

\subsection{Semantic-Spatial Fusion}
\label{sec:semantic-spatial-fusion}

We combine the two interface outputs via a simple linear fusion $\mathcal{F}$, treating the spatial score as a symmetric residual bias over the semantic base rather than introducing tunable interaction terms. We evaluate $T_{\mathrm{target}}$ against the frozen field to obtain $s_i^{\mathrm{sem}}$, min-max normalized over $K_{\mathrm{target}} = \{i \mid s_i^{\mathrm{sem}} > S_{th}\}$ (as in Sec.~\ref{sec:anchor-localization}) to match the $[-1,1]$ range of $s_i^{\mathrm{dir}}$:
\begin{equation}
\tilde{s}_i^{\mathrm{sem}} = 2\frac{s_i^{\mathrm{sem}} - \min_{j\in K_{target}} s_j^{\mathrm{sem}}}
{\max_{j\in K_{target}} s_j^{\mathrm{sem}} - \min_{j\in K_{target}} s_j^{\mathrm{sem}} + \epsilon} - 1.
\label{eq:semantic-normalization}
\end{equation}
The composite score is $m_i = \tilde{s}_i^{\mathrm{sem}} + \lambda s_i^{\mathrm{dir}}$, with $\lambda$ balancing spatial geometry against semantic confidence; sign-consistency of $s_i^{\mathrm{dir}}$ (Sec.~\ref{sec:direction-scoring}) lets this additive form boost aligned candidates and penalize misaligned ones directly.

Composite scores are splatted onto the target view $V$ following standard 3DGS rendering, producing a response map $M \in \mathbb{R}^{H \times W}$, binarized at pixel $\mathbf{p}$ as:
\begin{equation}
M_{\mathrm{binary}}(\mathbf{p}) = \mathbb{I} \big( M(\mathbf{p}) > \tau_{\mathrm{mask}} \big)
\label{eq:mask-binarization}
\end{equation}

\section{Experiment}
\label{sec:experiment}
\subsection{Experimental Setup}
\label{sec:experimental-setup}
\textbf{Dataset.} To evaluate observer-centric reasoning, we utilize the real-world Ramen and Waldo Kitchen scenes from Ref-LERF~\cite{he2025refersplat}. Because the original dataset lacks viewpoint-dependent relationships, we augmented these scenes with new, human-verified directional queries. For fair comparison, we include two official pre-trained scenes from the Ref-LERF dataset without retraining baseline models. Since DSSR-3D operates at inference time atop ReferSplat's frozen embeddings-the same embeddings used by the ReferSplat baseline-retraining them per scene would alter the input to both methods at once, confounding any gain with retraining variance rather than isolating our spatial reasoning contribution.

\begin{figure}[!t]
    \centering
    \includegraphics[width=0.65\linewidth]{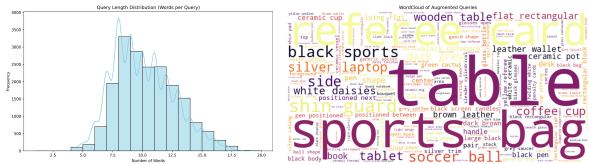}
    \caption{Word cloud and query length histogram of the evaluation prompts, illustrating the distribution of vocabulary usage and sentence lengths across all queries.}
    \label{fig:data_stat}
\end{figure}

We further propose ViewRef-GS, a new synthetic benchmark of four scenes (Sport, Sofa, Office, Room) reconstructed from 360-degree videos into 3D Gaussians following~\cite{hollein2026world}, providing over 700 calibrated views across 50 object instances. Combined with the two annotated Ref-LERF scenes, our full evaluation suite spans six scenes-two real-world (Ramen, Waldo Kitchen) and four synthetic-with over 19,000 queries uniformly covering 8 directional relations. As shown in Fig.~\ref{fig:data_stat}, query length concentrates between 8-12 words, and the vocabulary spans everyday object categories (furniture, kitchenware, personal items, sports equipment) qualified by material/color attributes and explicit relational phrases (e.g., ``positioned next''), reflecting the benchmark's focus on fine-grained, spatially grounded referring expressions. Since we generated ground-truth masks for ViewRef-GS using GroundingSAM~\cite{ren2024grounded}, we exclude it from quantitative comparison on these four scenes to avoid evaluating a method against ground truth it helped produce, restricting its use as a baseline to the Ref-LERF scenes, where its official annotations are independent of our pipeline.

\textbf{Implementation Details.} All experiments run on a single NVIDIA A100 GPU. For our main evaluations, we adopt ReferSplat~\cite{he2025refersplat} as the default frozen backbone, using official checkpoints for Ref-LERF and standard training protocols for ViewRef-GS. We apply a single fixed hyperparameter configuration across all six scenes: softmax temperature $\tau=0.5$, confidence floor $S_{\mathrm{th}}=0.7$ (viable across backbones, as all output normalized similarities), adaptive spatial decay $\sigma_d$ computed per query as the median distance of top semantic candidates from the anchor, and mask threshold $\tau_{\mathrm{mask}}=0.6$.

\textbf{Baselines.} We compare against representative methods spanning 3DGS-based referring segmentation, open-vocabulary 3D scene understanding, and 2D visual grounding: ReferSplat~\cite{he2025refersplat} (official checkpoints, default settings); LangSplat~\cite{qin2024langsplat} and GaussianGrouping~\cite{ye2024gaussiangrouping}, retrained on our benchmark scenes per their original protocols due to the absence of pre-trained models (GaussianGrouping, lacking native text-based referring, is seeded with a GroundingSAM~\cite{ren2024grounded} mask on the first frame and propagated across views); and GroundingSAM itself as a 2D baseline. Prior point-based methods~\cite{shi2024viewpointaware,nanri2026viewpoint3dseg} are excluded due to unavailable code and their dependence on 3D bounding-box annotations, which our bounding-box-free benchmarks do not provide.

\textbf{Evaluation Metric.} Following standard benchmarks for referring 3D segmentation, we use Intersection-over-Union (IoU) to measure localization accuracy. Given a predicted binary mask $M \in \{0,1\}^{H \times W}$, its ground-truth mask $G \in \{0,1\}^{H \times W}$, and the set of referring queries $\mathcal{Q}$, we report per-query IoU and mean IoU (mIoU) as:
\par\noindent
\makebox[\linewidth][c]{%
\begin{minipage}{0.42\linewidth}
\begin{equation}
\scalebox{0.82}{$\displaystyle
\text{IoU}(M,G)=\frac{|M\cap G|}{|M\cup G|}.
\vphantom{\frac{1}{|\mathcal{Q}|}\sum_{q\in\mathcal{Q}}\text{IoU}(M_q,G_q)}
$}
\label{eq:iou}
\end{equation}
\end{minipage}\hfill
\begin{minipage}{0.56\linewidth}
\begin{equation}
\scalebox{0.82}{$\displaystyle
\text{mIoU}=\frac{1}{|\mathcal{Q}|}\sum_{q\in\mathcal{Q}}\text{IoU}(M_q,G_q).
$}
\label{eq:miou}
\end{equation}
\end{minipage}
}
To evaluate view-dependent query augmentation, we compute mIoU over synthesized relational prompts that span diverse spatial constraints, camera viewpoints, and object configurations.

\subsection{Main Results}
\label{sec:main_results}

\begin{table*}[!t]
\centering
\caption{Quantitative comparison with 2D and 3D baselines (mIoU \%). ``Overall'' averages all 13-17 annotated objects per scene; selected columns are illustrative.}
\label{tab:main_results}
\resizebox{\textwidth}{!}{%
\begin{tabular}{cl cccc cccc}
\toprule
\multirow{2}{*}{\textbf{Type}} & \multirow{2}{*}{\textbf{Method}} & \multicolumn{4}{c}{\textbf{Ramen}} & \multicolumn{4}{c}{\textbf{Waldo Kitchen}} \\
\cmidrule(lr){3-6} \cmidrule(lr){7-10}
& & Glass & Chopsticks & Noodles & \textbf{Overall} & Plate & Yellow Desk & Red Cup & \textbf{Overall} \\
\midrule
2D & Grounding SAM & 27.22 & 30.00 & 23.17 & 27.73 & 0.13 & 0.00 & 47.28 & 19.42 \\
\midrule
\multirow{4}{*}{3D} 
& LangSplat  & 36.75 & 19.86 & 19.39 & 27.98 & 20.51 & 29.03 & 25.72 & 21.26 \\
& ReferSplat & 47.79 & 25.93 & 20.80 & 32.58 & 20.09 & 30.22 & 31.21 & 21.87 \\
& GS-Grouping & 29.77 & 34.08 & 33.47 & 34.43 & 20.66 & 24.51 & 20.00 & 21.43 \\
& \textbf{DSSR-3D} & \textbf{80.13} & \textbf{39.77} & \textbf{27.12} & \textbf{37.86} & \textbf{32.60} & \textbf{39.82} & \textbf{38.98} & \textbf{27.23} \\
\bottomrule
\end{tabular}%
}
\end{table*}

\begin{table}[!t]
\centering
\caption{Scene-level mIoU (\%) on ViewRef-GS. The \textcolor{green!50!black}{$\uparrow$} denotes our absolute improvement over ReferSplat.}
\label{tab:refersplat_epoch6_comparison}
\renewcommand{\arraystretch}{1.0}
\setlength{\tabcolsep}{4pt}
\resizebox{\linewidth}{!}{%
\begin{tabular}{lccccc}
\toprule
\textbf{Method} & \textbf{Sport} & \textbf{Sofa} &
\textbf{Office} & \textbf{Room} & \textbf{Average} \\
\midrule
LangSplat & 27.85 & 0.62 & 13.29 & 10.65 & 13.10 \\
ReferSplat & 26.99 & 50.78 & 16.19 & 49.15 & 35.78 \\
GS-Grouping & 17.06 & 36.63 & 8.86 & 23.19 & 21.44 \\
\midrule
\rowcolor{green!10}
\textbf{DSSR-3D} &
\textbf{37.43} \textcolor{green!50!black}{\scriptsize($\uparrow$ 10.44)} &
\textbf{57.15} \textcolor{green!50!black}{\scriptsize($\uparrow$ 6.37)} &
\textbf{22.62} \textcolor{green!50!black}{\scriptsize($\uparrow$ 6.43)} &
\textbf{52.15} \textcolor{green!50!black}{\scriptsize($\uparrow$ 3.00)} &
\textbf{42.34} \textcolor{green!50!black}{\scriptsize($\uparrow$ 6.56)} \\
\bottomrule
\end{tabular}
}
\end{table}


\textbf{Quantitative Analysis.} Table~\ref{tab:main_results} demonstrates the effectiveness of viewpoint-dependent spatial reasoning over existing 3DGS-based and 2D baselines: our method outperforms GS-Grouping, ReferSplat, and Grounding SAM on Ramen, and surpasses ReferSplat and GS-Grouping on Waldo Kitchen despite greater clutter and occlusion. Grounding SAM performs competitively on salient, single-viewpoint-visible objects such as Red Cup, but less consistently across the two scenes: its single-view formulation is limited to visible surface projections and therefore struggles with occluded targets, whereas our 3D-native representation can recover these targets from the scene geometry. Moreover, Grounding SAM requires a full forward pass for each query-viewpoint pair, while our approach amortizes the computation offline and supports efficient inference-time evaluation (Sec.~\ref{ab:time_cmp}).

Across the four synthetic scenes, our method consistently outperforms all 3DGS-based baselines (Table~\ref{tab:refersplat_epoch6_comparison}), with the largest absolute performance observed on Sofa. These gains stem from our Pose-Conditioned Directional Scoring interface (Sec.~\ref{sec:direction-scoring}), which explicitly conditions spatial reasoning on camera pose rather than relying on view-invariant semantic features, enabling the model to disambiguate directional relations across different scene layouts and observer viewpoints.


\textbf{Generality Across Semantic Backbones.} A core claim of our decoupled architecture (Sec.~\ref{sec:problem-formulation}) is that the spatial reasoning interfaces function independently of the underlying semantic field. While our main results adopt ReferSplat as the default backbone, we validate this modularity by applying the same training-free DSSR-3D module, without any per-backbone adaptation, atop LangSplat and GaussianGrouping. As shown in Table~\ref{tab:cross-backbone}, this consistently improves both across all six scenes: LangSplat rises from 16.94\% to 21.02\% average mIoU ($+4.08$), and GS-Grouping from 23.60\% to 28.43\% ($+4.83$), with gains holding on every individual scene rather than being driven by one environment.

This zero-shot transferability confirms that view-dependent spatial reasoning can be fully decoupled from semantic feature extraction-DSSR-3D is not a backbone-specific heuristic, but a generalized, plug-and-play layer applicable across 3DGS semantic fields.

\begin{table}[!t]
\centering
\caption{Cross-backbone generalization on all six scenes (mIoU \%). DSSR-3D is applied to each frozen semantic backbone without any retraining or hyperparameter tuning.}
\label{tab:cross-backbone}
\setlength{\tabcolsep}{7.5pt}
\resizebox{\linewidth}{!}{%
\begin{tabular}{@{}l ccc ccc c@{}}
\toprule
\multirow{2}{*}{\textbf{Architecture Setup}} & \multicolumn{2}{c}{\textbf{Real Scenes (Ref-LERF)}} & \multicolumn{4}{c}{\textbf{Synthetic Scenes (ViewRef-GS)}} & \multirow{2}{*}{\textbf{Mean}} \\
\cmidrule(lr){2-3} \cmidrule(lr){4-7}
& Ramen & Waldo Kitchen & Sport & Sofa & Office & Room & \\
\midrule
LangSplat~\cite{qin2024langsplat} & 27.98 & 21.26 & 27.85 & 0.62 & 13.29 & 10.65 & 16.94 \\
\quad \textbf{+ DSSR-3D (Ours)} & \textbf{30.74} & \textbf{21.94} & \textbf{32.18} & \textbf{5.86} & \textbf{16.47} & \textbf{18.92} & \textbf{21.02} {\scriptsize(\textbf{$+4.08$})} \\
\midrule
GS-Grouping~\cite{ye2024gaussiangrouping} & 34.43 & 21.43 & 17.06 & 36.63 & 8.86 & 23.19 & 23.60 \\
\quad \textbf{+ DSSR-3D (Ours)} & \textbf{36.21} & \textbf{22.38} & \textbf{24.67} & \textbf{42.54} & \textbf{13.72} & \textbf{31.08} & \textbf{28.43} {\scriptsize(\textbf{$+4.83$})} \\
\bottomrule
\end{tabular}
}
\end{table}

\begin{figure}[!t]
    \centering
    \includegraphics[width=0.90\linewidth]{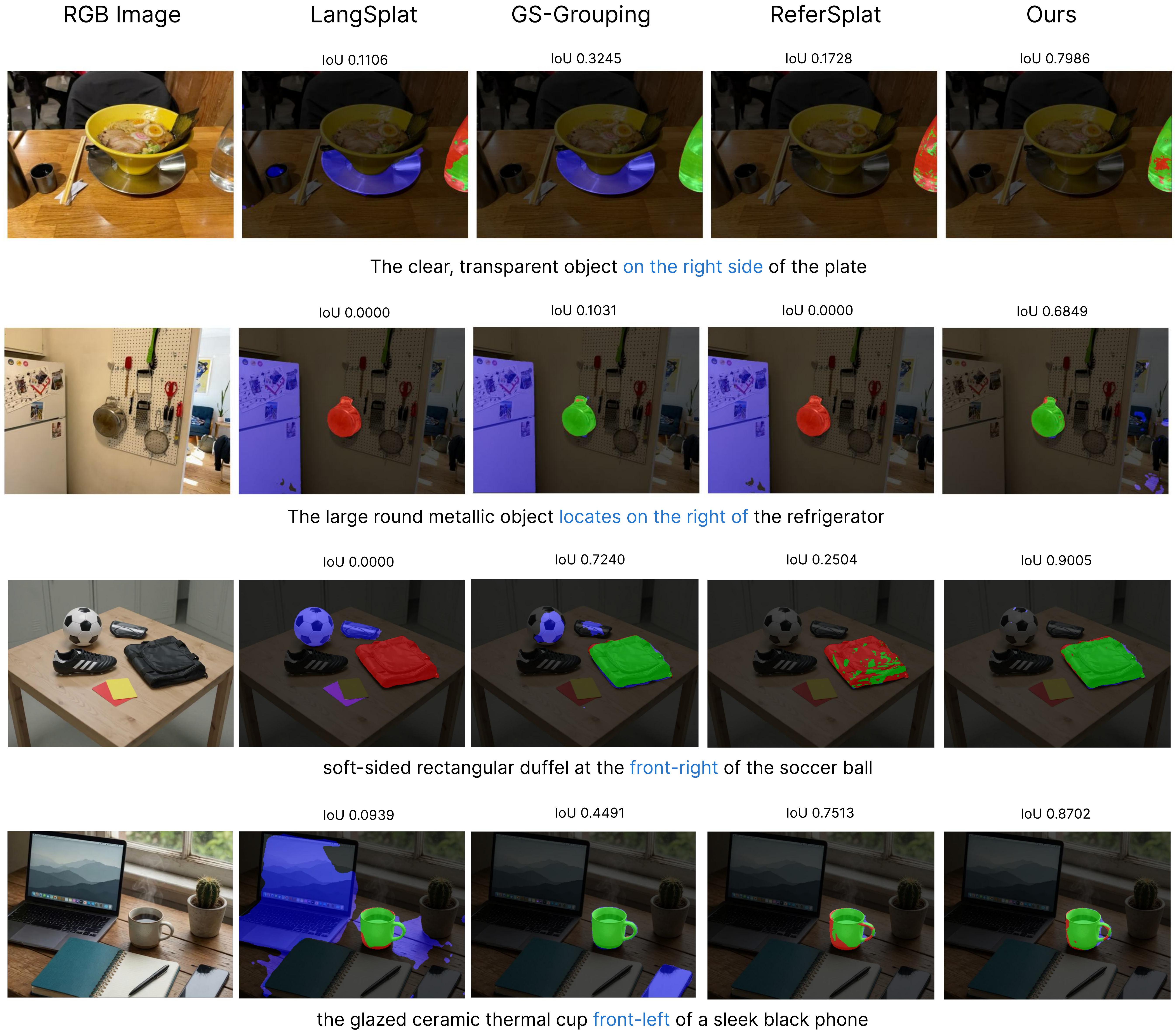}
    \caption{Qualitative comparison on augmented Ref-LERF and ViewRef-GS. Ground truth is \textbf{green}, false positives are \textbf{blue/purple}, and false negatives are \textbf{red}.}
    \label{fig:qualitative}
\end{figure}

\textbf{Qualitative Results.}
Figure~\ref{fig:qualitative} further examines our method's behavior. Compared to baselines, DSSR-3D generally produces more spatially consistent masks with improved boundary alignment and fewer missed regions, though false positives persist under occlusion or visually similar distractors. Notably, GS-Grouping's performance depends heavily on the initial view and GroundingSAM quality: early false positives in the first-frame grounding can propagate into incorrect Gaussian groupings across subsequent views.

\section{Ablation Study}
\label{sec:ablation-study}
\subsection{Time Comparison with Grounding SAM}
\label{ab:time_cmp}
\begin{table}[!t]
\centering
\caption{Total inference latency comparison (seconds per query), measured on a single NVIDIA A100 GPU.}
\label{tab:time_comparison}
\begin{tabular}{lcc}
\toprule
\textbf{Method} & \textbf{Ramen} & \textbf{Waldo Kitchen} \\
\midrule
Grounding SAM & $0.577$ & $0.697$ \\
\textbf{DSSR-3D} & \textbf{0.072} & \textbf{0.085} \\
\bottomrule
\end{tabular}
\end{table}

Table~\ref{tab:time_comparison} compares the inference latency of our framework with Grounding SAM, targeting the repeated online querying scenario where the one-time cost of constructing the 3D semantic field is amortized and excluded, while GroundingSAM pays its full cost on every query. Our reported runtime includes the full pipeline (parsing, localization, scoring, rasterization). GroundingSAM must repeatedly forward-pass heavy visual backbones per query-viewpoint pair, yielding $\sim$0.6-0.7s latency; our framework instead amortizes feature extraction offline, evaluating spatial reasoning directly within the pre-embedded Gaussian space via efficient rasterization, reducing online latency to $\sim$0.08s per query-an $\sim$8$\times$ speedup for repeated queries within a known scene.

\subsection{Anchor Localization Strategies}
\label{ab:anchor_strat}
To evaluate the effectiveness of the Temperature-Sharpened Softmax, we compare it with alternative anchor localization strategies. Ground-truth 3D bounding boxes would provide an upper bound, but neither Ref-LERF nor our generated dataset includes them, making such evaluation impractical without additional annotation. We instead adopt a bounding-box-free baseline based on hard assignment, selecting the top-$K$ Gaussians with the highest semantic responses and computing their unweighted centroid-a choice sensitive to $K$, trading off spatial coverage against noise inclusion: as shown in Table~\ref{tab:ablation_center_of_mass}, small $K$ limits coverage in larger scenes while large $K$ absorbs more background noise in cluttered ones.

Our Temperature-Sharpened Softmax relaxes this fixed-capacity constraint by weighting contributions according to semantic confidence, reducing the influence of low-confidence points while adapting to varying object extents. As reported in Table~\ref{tab:ablation_center_of_mass}, this formulation achieves the most consistent performance across all scenes, outperforming the strongest Top-$K$ result by 3.11 and 4.02 mIoU points on \textit{Ramen} and \textit{Waldo Kitchen}, respectively, while retaining its gains on the synthetic scenes.
\begin{table}[!t]
\centering
\caption{Anchor-localization ablation across six scenes (mIoU \%): static Top-$K$ assignment versus Temperature-Sharpened Softmax.}
\label{tab:ablation_center_of_mass}
\setlength{\tabcolsep}{3pt}
\resizebox{\linewidth}{!}{%
\begin{tabular}{lcccccc}
\toprule
\textbf{Method} & \textbf{Ramen} & \textbf{Waldo Kitchen} & \textbf{Room} & \textbf{Office} & \textbf{Sofa} & \textbf{Sport} \\
\midrule
Top-$K$ ($K=50$)   & 34.21 & 23.21 & 30.54 & 13.29 & 46.56 & 24.05 \\
Top-$K$ ($K=200$)  & 34.28 & 23.21 & 42.06 & 15.93 & 49.52 & 26.07 \\
Top-$K$ ($K=800$)  & 34.75 & 22.93 & 49.15 & 16.49 & 52.49 & 28.93 \\
\midrule
\textbf{Softmax (DSSR-3D)} & \textbf{37.86} & \textbf{27.23} & \textbf{52.15} & \textbf{22.62} & \textbf{57.15} & \textbf{37.43} \\
\bottomrule
\end{tabular}
}
\vspace{-3mm}
\end{table}



\subsection{Robustness of Spatial Decay}
\label{sec:ablation_spatial_decay}

We ablate the spatial scale parameter $\sigma_d$ (Eq.~\ref{eq:spatial-decay}), comparing an \textit{adaptive} variant defined per query as the median distance of top semantic candidates from the anchor against fixed-radius configurations to address the brittleness of fixed thresholds across varying object sizes. As shown in Table~\ref{tab:spatial_decay}, Adaptive $\sigma_d$ consistently outperforms both the unpenalized baseline and all fixed radii ($\sigma_d \in \{0.5, 1.0, 2.0\}$) across all six scenes. Fixed values yield limited, inconsistent gains, with the best-performing radius varying unpredictably by scene (e.g., $\sigma_d=2.0$ best in Waldo Kitchen, $\sigma_d=1.0$ best in Sofa), indicating no single radius generalizes well. Adaptive $\sigma_d$ instead wins on every scene, improving over the unpenalized baseline by up to 12.31 points (Sport) and at least 3.50 points (Ramen)-confirming that scaling the decay radius to each query's local semantic distribution, rather than fixing it a priori, is necessary for consistent performance. We adopt it as the default for all main results (Sec.~\ref{sec:main_results}).

\begin{table}[!t]
\centering
\caption{Spatial-decay ablation across six scenes (mIoU \%). Adaptive $\sigma_d$ derives the penalty radius from semantic candidates.}
\label{tab:spatial_decay}
\setlength{\tabcolsep}{3pt}
\resizebox{0.90\linewidth}{!}{%
\begin{tabular}{lcccccc}
\toprule
\textbf{Configuration} & \textbf{Ramen} & \textbf{Waldo Kitchen} & \textbf{Room} & \textbf{Office} & \textbf{Sofa} & \textbf{Sport} \\
\midrule
w/o Spatial Decay    & 34.36 & 22.49 & 49.32 & 16.15 & 47.18 & 25.12 \\
$\sigma_d = 0.5$     & 33.58 & 22.56 & 49.52 & 16.15 & 47.09 & 25.15 \\
$\sigma_d = 1.0$     & 33.71 & 22.63 & 49.46 & 16.09 & 47.90 & 25.52 \\
$\sigma_d = 2.0$     & 34.06 & 22.74 & 49.52 & 16.13 & 47.47 & 25.28 \\
Adaptive $\sigma_d$  & \textbf{37.86} & \textbf{27.23} & \textbf{52.15} & \textbf{22.62} & \textbf{57.15} & \textbf{37.43} \\
\bottomrule
\end{tabular}
}
\vspace{-3mm}
\end{table}

\section{Conclusion}
\label{sec:conclusion}
We propose \textbf{DSSR-3D}, an inference-time framework for view-dependent referring segmentation that decouples view-invariant semantic localization from view-conditioned spatial reasoning. Because our training-free interfaces generalize across structurally distinct baselines without adaptation, this absence of backbone-specific learning is a design advantage, not a limitation. Instantiated via soft continuous aggregation and algebraic projection, we validate this across six scenes and eight directional relations on Ref-LERF and our new \textbf{ViewRef-GS} benchmark.

\textbf{Future work.} Our formulation trades exact boundary precision for a bounding-box-free, continuous representation, which may introduce directional bias for elongated or topologically complex objects; the pipeline is also scoped to single-anchor queries. Future work will extend to multi-anchor relationships and larger-scale, more diverse benchmarks.

\bibliographystyle{plain}
\nocite{*}
\bibliography{reference}

\end{document}